\documentclass[sigconf]{acmart}

\usepackage{graphicx}
\AtBeginDocument{%
  }

\setcopyright{acmlicensed}
\copyrightyear{2018}
\acmYear{2018}
\acmDOI{XXXXXXX.XXXXXXX}

\acmConference[Conference acronym 'XX]{Make sure to enter the correct
  conference title from your rights confirmation emai}{June 03--05,
  2018}{Woodstock, NY}
\acmISBN{978-1-4503-XXXX-X/18/06}

\begin{document}

\title{AI in Manufacturing: Market Analysis and Opportunities}

\author{Mohamed Abdelaal}
\email{mohamed.abdelaal@softwareag.com}
\affiliation{%
  \institution{Software AG, Darmstadt, Hessen, Germany}
  \city{}
  \state{}
  \country{}
}

\renewcommand{\shortauthors}{Abdelaal et al.}

\begin{abstract}
In this paper, we explore the transformative impact of Artificial Intelligence (AI) in the manufacturing sector, highlighting its potential to revolutionize industry practices and enhance operational efficiency. We delve into various applications of AI in manufacturing, with a particular emphasis on human-machine interfaces (HMI) and AI-powered milling machines, showcasing how these technologies contribute to more intuitive operations and precision in production processes.

Through rigorous market analysis, the paper presents insightful data on AI adoption rates among German manufacturers, comparing these figures with global trends and exploring the specific uses of AI in production, maintenance, customer service, and more. In addition, the paper examines the emerging field of Generative AI and the potential applications of large language models in manufacturing processes. The findings indicate a significant increase in AI adoption from 6\% in 2020 to 13.3\% in 2023 among German companies, with a projection of substantial economic impact by 2030. The study also addresses the challenges faced by companies, such as data quality and integration hurdles, providing a balanced view of the opportunities and obstacles in AI implementation.
%
\end{abstract}
\maketitle

\section{Introduction}\label{sec:introduction}

Manufacturing has been one of the early adopters of artificial intelligence (AI) and has witnessed significant benefits from the technology's implementation. With the use of AI, manufacturers can analyze vast amounts of data generated by sensors, machines, and other sources to identify inefficiencies, predict maintenance needs, and optimize production schedules. Additionally, AI can broadly improve safety in manufacturing environments by analyzing data in real-time and identifying potential safety hazards before they cause accidents. By adopting AI technologies, manufacturing companies can gain a competitive advantage in the market and stay ahead of the curve in an ever-changing business landscape. In this report, we explore the potential of AI in manufacturing, with a particular focus on human-machine interfaces (HMI) and AI applications for milling machines. 

The report identifies human-machine interfaces as a key area where AI is making a significant impact in the manufacturing industry. AI is helping to create more intuitive and user-friendly interfaces that make it easier for operators to control and monitor manufacturing processes. Moreover, the report highlights the use of AI in milling machines. By leveraging AI, milling machines can make real-time adjustments to their operations, leading to more accurate and precise machining, and ultimately, better products. The report provides insights into how AI is transforming the manufacturing industry and how companies can benefit from implementing AI in their operations. In this report, we provide a market analysis of AI applications in manufacturing. Specifically, we provide a list of use cases related to human-machine interaction and the use of AI for milling machines. We also answer some questions such as how many companies in Germany use AI? How many of those companies are in manufacturing? Which AI application is the most common in manufacturing? 

In summary, this article makes significant contributions to the understanding of AI adoption and its potential in the manufacturing industry. By providing a comprehensive market analysis, the report offers valuable insights into the current state of AI implementation, the challenges faced by manufacturers, and the successful use cases that demonstrate the benefits of AI-driven solutions. We break down the manufacturing process into its constituent phases, identifying key areas where AI can be integrated to optimize performance and efficiency. Additionally, we explore the role of human-machine interfaces and the potential of Generative AI and large language models in revolutionizing various aspects of manufacturing. By presenting a detailed examination of AI's transformative potential, this article serves as an essential resource for stakeholders in the manufacturing sector, guiding them toward a more competitive and innovative future.

The remainder of the article is structured as follows. Section~\ref{sec:facts} delves into the adoption of AI by German manufacturers, presenting survey data from various sources and discussing the challenges faced by companies in implementing AI technologies. Section~\ref{sec:manufacturing} provides an overview of the typical phases of the manufacturing process, setting the stage for the subsequent discussion of AI use cases. Section~\ref{sec:use_cases} forms the core of the article, presenting a comprehensive list of AI use cases in manufacturing, supported by real-world examples of German companies successfully implementing these solutions. Section~\ref{sec:ai_hmi} is dedicated to the role of AI in enhancing human-machine interfaces, before Section~\ref{sec:milling} introduces the potential of AI for milling machines. The potential of Generative AI and large language models in manufacturing is explained in Section~\ref{sec:genai}. Finally, Section~\ref{sec:conclusion} concludes the article with a summary of its contributions.

\section{AI-Adoption by German Manufacturers}\label{sec:facts}

\begin{figure}
    \centering
    \includegraphics[width=\columnwidth]{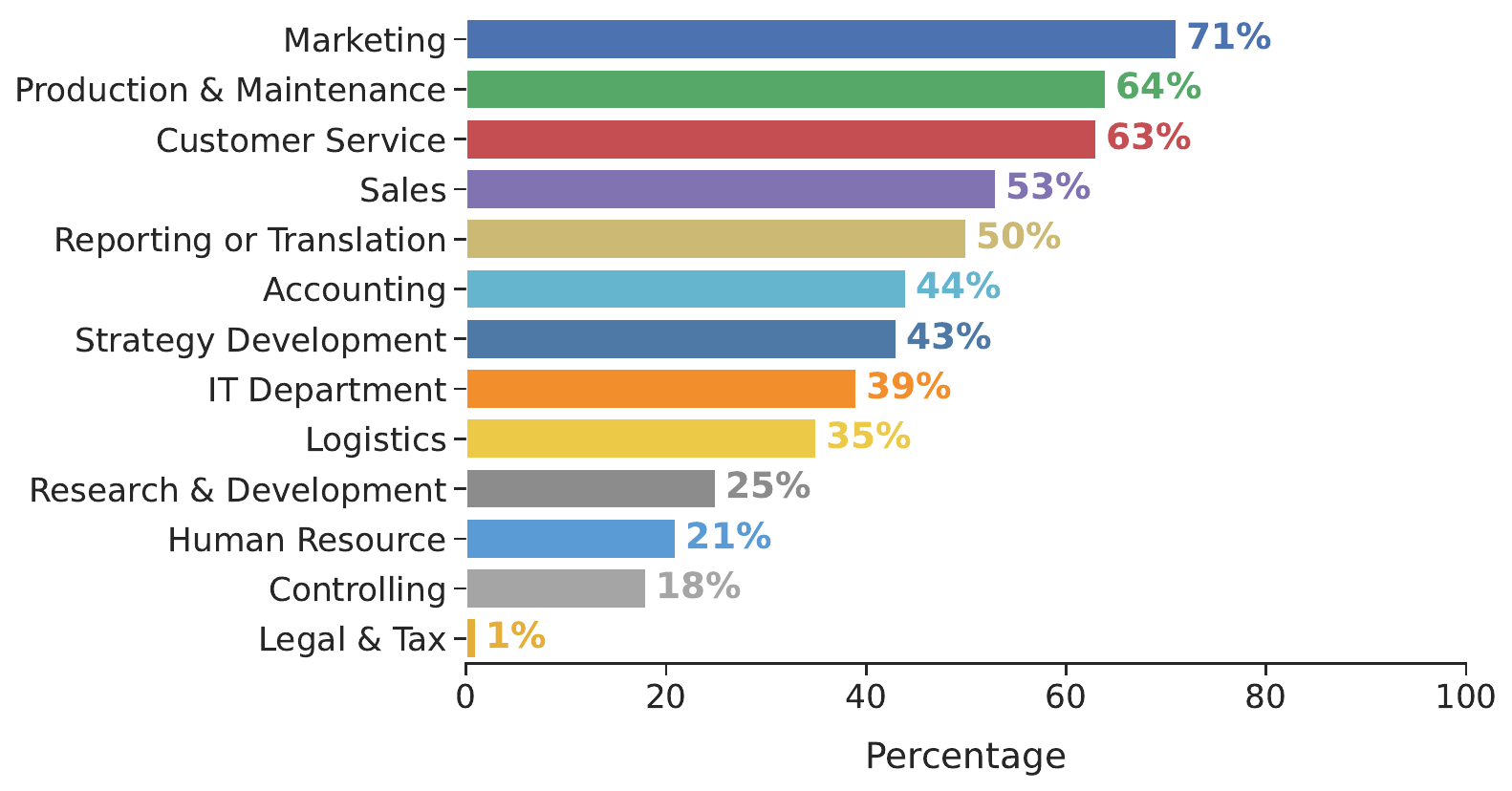}
    \caption{AI adoption in German companies (adapted from~\cite{StripedGiraffe2021})}
    \Description{A bar chart showing the percentage of AI adoption in various sectors of German companies. The chart highlights significant adoption in the marketing and manufacturing sectors.}
    \label{fig:ai_germany}
\end{figure}

In this section, we explore the various studies of AI involvement in German firms, putting more focus on the manufacturing sector. There exist several surveys and reports that investigated the adoption of AI in German manufacturing. For instance, Statista recently released a report about AI adoption in Germany \cite{statista2024ai}. The report states that the AI market is projected to reach a size of \$7.85 billion in 2024. It is expected to grow at an annual rate (CAGR) of 28.41\% from 2024 to 2030, resulting in a market volume of \$35.19 billion by 2030. According to the ifo Business survey in 2023, 13.3\% of German companies are currently using AI, and 9.2\% are planning to adopt it. Additionally, 36.7\% of all companies surveyed are exploring potential use cases for AI \cite{ifo2023ai}. Such results show a better position of AI adoption than the survey conducted by the digital association Bitkom in 2021 which states that AI applications are used in only 8\% of German companies, compared to 6\% in 2020 \cite{StripedGiraffe2021}. Nevertheless, nearly 70\% of German companies believe AI is a key technology of the future. The Bitkom's survey also examined what AI is used for in the German companies that have already implemented this technology. German companies predominantly use AI for marketing purposes, with personalized advertising being the most common application (71\%). Furthermore, 64\% of these businesses utilize AI to optimize their internal production and maintenance processes. Approximately 63\% of companies integrate AI into their customer service, utilizing it to automatically answer queries. Additionally, AI is employed by about 53\% of companies to analyze customer behavior in sales and by 50\% of companies to prepare text-based materials such as reports or translations across various departments. Figure~\ref{fig:ai_germany} delineates the distributions of AI applications adopted by German companies. The figure highlights that the highest percentage of companies using AI is in marketing, followed by production and maintenance, customer service, and sales.

Focusing on production, we found that the results obtained in Bitkom’s survey broadly conform with the results obtained in another survey by Capgemini in 2019. Specifically, the survey carried out by Capgemini showed that 69\% of the manufacturers in Germany use AI in their day-to-day operations, compared to 28\% in the US and 11\% in China~\cite{Capgemini2019}. By 2030, German companies in the production domain are expected to generate an additional 430 billion euros from AI applications. As a result, the gross domestic product at the end of the decade will be 11.3 percent higher than it would have been without using this technology, according to a study by the consulting firm PricewaterhouseCoopers \cite{mischler2019ki}. The survey by Capgemini demonstrated that AI applications used at German companies in the manufacturing sector include process and logistics data analysis, quality assurance, machine control, data-driven process modeling, online condition monitoring, anomaly detection, adaptive control systems, and simulation technology. 

Aside from the BitKom's survey, there exist also other surveys with distinct conclusions. For instance, 
the survey by Statista in 2019 found that 23\% of German industrial companies with 500 to 1,999 employees used AI in the context of Industry 4.0 practices~\cite{Koptyug2019}. A more recent survey conducted by Harris Poll on behalf of Google Cloud in 2021 found that 79\% of manufacturers in Germany are leveraging AI on a daily basis, with quality control and supply chain optimization being the two main application areas for AI in manufacturing~\cite{Wee2021}. Such a survey revealed that the top three manufacturing sub-sectors deploying AI to assist in day-to-day operations are automotive/OEMs with 76\%, automotive suppliers with 68\%, and heavy machinery with 67\%. Finally, a recent survey conducted by ZEW in 2023 revealed that only 10\% of manufacturing companies and 30\% of companies in the information economy in Germany had implemented AI \cite{zew2024concerns}. Additionally, another quarter of companies in both sectors planned to adopt AI in the future. However, only a small percentage of companies reported having a high level of competitiveness in AI: only 6\% in the manufacturing industry and 15\% in the information economy.

\begin{figure}
    \centering
    \includegraphics[width=\columnwidth]{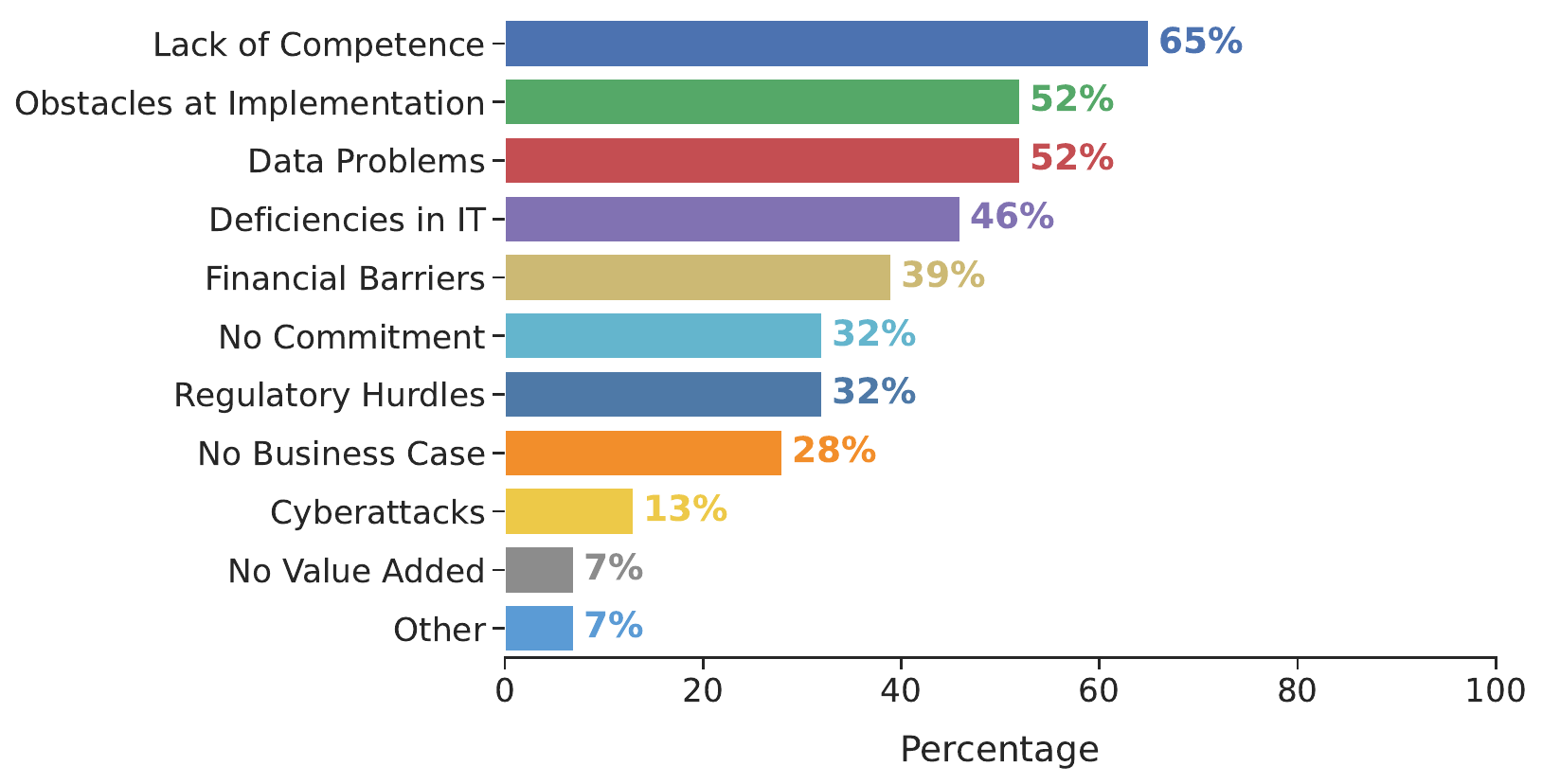}
    \caption{Barriers to AI in German SMEs (adapted from~\cite{Ulrich2021})}
    \Description{A bar chart showing the barriers to AI adoption in German SMEs. The chart highlights the lack of competencies and data problems as the main barriers.}
    \label{fig:barriers}
\end{figure}

Despite its potential, German manufacturers often encounter numerous challenges in the successful and effective implementation of AI in their production environments. One of the main challenges is the need for clean, meaningful, high-quality data, which is critical for the success of AI initiatives but can be a challenge in manufacturing. Biases in manufacturing data can be caused by multiple factors, such as unrepresentative or incomplete training data or the reliance on flawed information. For example, collecting data from a specific group of workers may lead to biased data that does not reflect the entire workforce. Additionally, biases can occur when data is collected from a specific time period, leading to outdated data that does not reflect current conditions. Other challenges stem from the lack of business alignment, difficulty assessing vendors, integration challenges, and legal issues. Furthermore, Germany lags behind other countries in developing AI applications, which may make it more difficult for German manufacturers to implement AI in their operations~\cite{Kinkartz2019}. Figure~\ref{fig:barriers} demonstrates a set of barriers to AI in German small and medium enterprises (SMEs), which have been identified in an empirical study in 2021~\cite{Ulrich2021}. 

\section{Manufacturing Process}\label{sec:manufacturing}

\begin{figure}
    \centering
    \includegraphics[width=1\columnwidth]{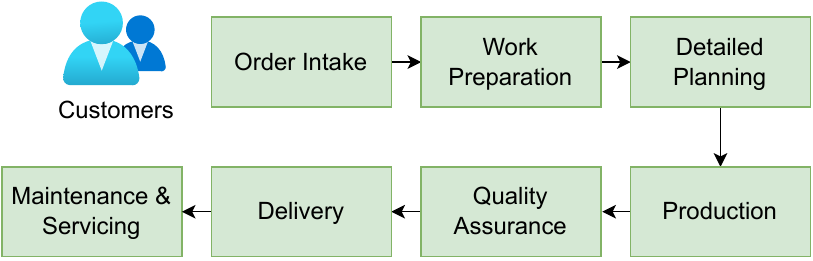}
    \caption{Typical phases of the manufacturing process}
    \label{fig:manufacturing}
\end{figure}

In this section, we elaborate on the AI use cases in manufacturing. Before delving into the use cases and for the report to be self-contained, we provide a glimpse into the typical phases of the manufacturing process. Subsequently, we explore the possibility of integrating AI into the production process, providing a list of specific use cases for AI in each phase. In general, the manufacturing process can be interpreted as a chronological sequence of several activities, including sourcing planning, resource management, personnel planning, and optimization of processes in inbound and outbound logistics. Figure~\ref{fig:manufacturing} depicts a set of typical phases of the manufacturing process. 

The manufacturing process typically begins with the order intake when the company receives the customer's request for a specific product or service. The order intake process involves several steps, including capturing the customer's requirements, determining the feasibility of the project, and providing the customer with a quote or estimate. During this phase, the company must establish clear communication with the customer to ensure they fully understand their requirements and are satisfied with the proposed solution. The accuracy and completeness of the order intake process can impact the rest of the production process and ultimately affect customer satisfaction. Therefore, it is essential to carefully manage this phase to ensure that all customer requirements are met and properly communicated to the production team. 

In the work preparation phase, the production team analyzes the order details and plans how to complete the project. Resource management, supply chains, and scheduling are the key areas to focus on during this phase. The company must ensure that the necessary resources, including personnel, materials, and equipment, are available to complete the project. The company must also consider the supply chain, ensuring that all necessary components and materials are available when needed to avoid production delays. The scheduling of the project is critical to ensure timely delivery, and it involves coordinating various production activities to ensure that everything happens according to plan. The work preparation phase lays the foundation for the success of the entire production process.

In the detailed planning phase, the project plan is finalized. This involves final time management, commissioning, and other activities necessary to get the project ready for production. The company must ensure that all steps in the production process are identified and scheduled, including quality control checks and any necessary equipment maintenance. The commissioning process is essential in ensuring that the final product meets all requirements before delivery. During this phase, the company must ensure that all stakeholders are aware of the project plan and timelines to ensure everyone is on the same page. The production phase is where the actual manufacturing and assembly of the product take place. It includes transporting the necessary materials and components to the production site and assembling them according to the project plan. Post-work or improvement is also carried out during this phase to ensure that the product meets all quality requirements. The company must ensure that all production activities are carefully monitored to avoid any quality issues that can arise during the manufacturing process.

The quality assurance phase is where the company carries out a series of quality checks on the product to ensure it meets all quality requirements. This includes checking for defects, ensuring that the product meets all specifications, and conducting any necessary tests. The company must ensure that all quality checks are carried out according to the quality control plan to avoid any issues that can affect the final product's quality. The delivery phase involves shipping the product to the customer or storing it for future use. This involves coordinating with the shipping company to ensure that the product is delivered according to the customer's specifications. The company must also ensure that the product is stored appropriately if it is not immediately required by the customer. This phase is critical to ensuring that the customer receives the product on time and in excellent condition. The maintenance phase is the final stage of the production process. This involves maintaining the product, ensuring that it is functioning correctly, and addressing any issues that arise. The company must ensure that all maintenance activities are carried out according to the maintenance plan, including any necessary repairs or upgrades. The maintenance phase is critical to ensuring that the product continues to meet the customer's requirements and operates efficiently.
\section{AI Use Cases in Manufacturing}\label{sec:use_cases}

In this section, we discuss the potential of AI in the manufacturing process, with providing examples of real-world use cases of the AI adoption by German manufacturers. 

\paragraph{Chatbots} In fact, AI-powered chatbots can be used to interact with customers and gather their order details, which can reduce the workload on human employees and free them up to focus on other important tasks. The chatbots are not only used as a communication tool with the customers but they can also be leveraged as an effective tool for facilitating repeated tasks done by the employees. As an example, the Avanade chatbot adopted by Siemens~\cite{Avanada2023} has improved the overall digital workplace experience for Siemens employees. It saves them time searching for up-to-date information and getting answers to repetitive questions.

\paragraph{Digital Twins} In general, digital twins are virtual representations of physical objects or systems that span their lifecycle, are updated from real-time data, and use simulation. The integration of AI and machine learning algorithms with digital twins allows businesses to quickly analyze and interpret data to make informed decisions~\cite{Lv2022}. This integration of AI improves digital twin technology by enabling businesses to obtain granular insights into their operations, enabling them to optimize performance, reduce downtime, and improve product quality. AI algorithms can be used to analyze large amounts of data generated by digital twins to identify patterns, predict outcomes, and optimize performance.  In 2020, the worldwide digital twins market was worth \$3.1 billion. However, it is projected to surpass \$48.2 billion by 2026~\cite{Spatial2022}.

German manufacturers have been at the forefront of digital twin technology and AI integration. Siemens, for example, has developed a digital twin of a gas turbine that uses AI algorithms to optimize its performance~\cite{Kerris2021}. Bosch has developed a digital twin of a production line that uses AI to predict maintenance needs and reduce downtime. Volkswagen VW and Bosch are using data from the Golf 8 in Europe and a digital twin model to improve the performance of self-driving cars by simulating the performance of the car in different driving conditions~\cite{flaherty2021bosch}. The digital twin of the BMW factory replicates every aspect of the production process, including avatars of workers transporting equipment along the production lines, technicians shaping the metal frame of each vehicle, and quality control experts thoroughly inspecting each component~\cite{Spatial2022}.

\paragraph{Data Analysis}
AI algorithms can analyze customer data and purchase history to make personalized recommendations, which can increase the chances of customers making additional purchases. Additionally, AI can help automate the order-taking process by automatically processing orders and generating invoices, reducing the time and effort required to complete these tasks manually~\cite{agiang2021intelligent}. For instance, Thyssenkrupp Materials Services has developed an AI-powered platform called ``alfred'' that uses ML algorithms to analyze customer data and make personalized recommendations for each customer~\cite{tubbesing2019thyssenkrupp}. The platform has helped to increase customer loyalty and improve the efficiency of the order-taking process. 

\paragraph{Supply Chain Optimization}
AI can help manufacturers optimize their supply chains by analyzing data on suppliers, inventory levels, and customer demand. For example, AI algorithms can analyze historical data on supplier performance and delivery times to identify potential bottlenecks and proactively manage inventory levels. This can help to reduce lead times and improve customer satisfaction. Moreover, AI-enabled systems for supply chain management can help manufacturers assess various scenarios to improve last-mile deliveries, predict optimal delivery routes, track driver performance in real-time, and assess weather and traffic reports besides historical data to forecast future delivery times accurately.  McKinsey predicts that AI-enhanced supply chains will reduce forecasting errors by 20-50\%, reduce lost sales by 65\%, and reduce the over-stocking inventories by 20-50\%~\cite{mewari2021ai}. As an example, BMW has implemented an AI-powered supply chain optimization system that uses ML algorithms to analyze data on suppliers, inventory levels, and customer demand~\cite{diianni2021bmw}. The system has helped BMW to broadly reduce inventory costs and improve on-time delivery rates. Another example is that of the carmaker Rolls Royce, which leverages advanced ML algorithms and image recognition to power its fleet of self-driving ships, which in turn improves its supply chain efficiency and safely transports its cargo~\cite{mewari2021ai}.

\paragraph{Predictive Maintenance} 
ML algorithms for predictive maintenance can help manufacturers identify potential equipment failures before they occur. To this end, AI-driven technologies have the capability to gather and analyze vast amounts of data, such as audio, video, and GPS, obtained from sensors situated on the factory floor. These technologies can identify abnormalities or inefficiencies in equipment, which can forestall unexpected machinery breakdowns. Such anomalies could take the form of unusual sounds emanating from a vehicle's engine or a malfunction in an assembly line. Accordingly, AI technologies can enhance factory efficiency and reduce costs by avoiding unscheduled equipment downtime. According to a report by McKinsey, the most significant benefit of AI in the manufacturing industry is through predictive maintenance, which has the potential to generate global value between $0.5 to $0.7 trillion~\cite{mewari2021ai}. BCG has labeled predictive maintenance as the primary Industry 4.0 focus, with particular importance for cement manufacturers.    

\paragraph{Quality Control}
A system capable of detecting defects that may go unnoticed by the human eye and triggering corrective measures automatically can broadly help reduce product recalls and minimize wastage. Additionally, the system can detect anomalies such as toxic gas emissions in real time, which can prevent workplace hazards and enhance worker safety in factories. In this context, computer vision for quality control can help manufacturers detect defects and anomalies in products and processes. For example, the BMW Group uses automated image recognition for quality checks, and inspections, and to eliminate pseudo-defects (deviations from target despite no actual faults)~\cite{mewari2021ai}. As a result, they have achieved high levels of precision in manufacturing. Another company that’s benefited from AI in manufacturing is Porsche. They use autonomous guided vehicles (AGVs) to automate significant portions of automotive manufacturing. The AGVs take vehicle body parts from one processing station to the next, eliminating the need for human intervention and making the facility resilient to disruptions like pandemics.

\paragraph{Factory Automation}
By leveraging AI-driven process mining tools, manufacturing companies can automatically identify and remove production process bottlenecks, while also facilitating performance comparison across different regions. This, in turn, enables them to standardize and streamline workflows to create better manufacturing processes. Another use case involves robotic process automation, where robots independently execute repetitive tasks on the factory floor, with human intervention being necessary only in case of exceptions or anomalies. Additionally, robots can employ computer vision to screen and examine processes without any human intervention. Accordingly, robotics and automation for factory operations can help manufacturers increase efficiency, reduce costs, and improve safety.  For instance, McKinsey suggests that the implementation of AI in the semiconductor industry for process automation can increase yields by up to 30\%, while also reducing testing costs and scrap rates~\cite{mewari2021ai}.

\paragraph{AI-Powered Design}
AI-enabled software can help create several optimized designs for a single product. The software, also known as generative design software, requires engineers to provide certain input parameters such as raw materials, size and weight, manufacturing methods, and cost and other resource constraints. Using these parameters, the algorithm can generate various design permutations. The software lets engineers test various designs against a wide collection of manufacturing scenarios and conditions to pick the best possible outcome. The carmaker Nissan is using AI to develop never-seen-before car designs in the blink of an eye. The process would take human designers months, or even years to complete. Such software can also be used to pick the right recipes that lead to the least amount of raw material and energy waste~\cite{mewari2021ai}.

\paragraph{Resource Management}
AI can help manufacturing firms manage their resources more efficiently by analyzing historical data on resource utilization and predicting future demand. For example, AI algorithms can analyze data on machine usage and downtime to optimize scheduling and reduce idle time. This can help to improve production efficiency and reduce costs. Since they collect data in real-time, manufacturers can monitor their warehouses continuously and plan their logistics better. Demand forecasting can further help manufacturers take action to stock up their warehouses in advance and keep up with the customer demand without enormous transportation costs. Robots in the warehouses can track, lift, move, and sort items, leaving the more strategic tasks to the humans and reducing workplace injuries. Automated quality control and inventorying can reduce warehouse management costs, improve productivity, and require a smaller labor force. As a result, manufacturers can increase their sales and profit margins~\cite{Kerris2021}. AI can help manufacturing firms optimize their production schedules by analyzing data on machine usage, labor availability, and customer demand. For example, AI algorithms can analyze historical data on production times and machine availability to create more accurate production schedules. This can help to reduce lead times and improve on-time delivery rates.

\section{Human-machine Interfaces}\label{sec:ai_hmi}

In this section, we introduce the human-machine interfaces and how they work. Subsequently, we discuss several ideas for the application of AI technologies to optimize the performance of human-machine interfaces. 

\subsection{What are Human Machine Interfaces?}
In general, human-machine interfaces (HMIs) serves as a mediator between a machine and the operating personnel, where it connects an operator to the controller for an industrial system. They typically include visual representations of the systems. Through an HMI, an operator can control the operation of machinery and associated devices in industrial environments. There are different types of Human Machine Interface (HMI) available in the market. The most common types of HMI are push-button, overseer, and data-handling HMIs~\cite{cope2022hmi}. Push-button HMIs are used to control simple machines and processes, and they are usually found on small machines. Overseer HMIs are used to monitor and control complex machines and processes, and they are usually found on large machines. Data handling HMIs are used to collect and analyze data from machines and processes, and they are usually found on machines that require constant feedback from the system or printed production reports.

Another way to categorize HMIs is based on their level of control. There are two basic types of HMI: supervisory level and machine level~\cite{carotron2020hmi}. On the one hand, supervisory-level HMIs are usually designed for control room environments and used for system-wide monitoring and control. On the other hand, machine-level HMIs use embedded, machine-level devices within the production facility itself. In addition to these types, there are also touch-screen HMIs, mobile HMIs, and web-based HMIs. Touch screen HMIs are used to control and monitor machines and processes through a touch screen interface. Mobile HMIs are used to control and monitor machines and processes through a mobile device. Web-based HMIs are used to control and monitor machines and processes through a web browser.

\subsection{AI-Adoption to HMI Systems}
In general, HMIs can allow humans to interact with machines using natural language, gestures, and other intuitive methods, and AI can interpret and respond to these inputs. There exist currently several examples of companies that are using HMIs to create new business models, such as healthcare companies using virtual assistants to provide medical advice and financial companies using chatbots to assist customers with banking transactions~\cite{gonfalonieri2020ai}. In this way, HMIs can improve efficiency, reduce costs, and increase customer satisfaction. Therefore, many companies nowadays explore ways to incorporate this technology into their operations. 

Traditionally, HMI solutions used to be independent, separate terminals that were installed by an OEM (Original Equipment Manufacturer) as a component of a machine. However, the latest HMI solutions that utilize Machine Learning have been designed to automatically transmit data to either an on-premise solution or the cloud, without any need for configuration. Nowadays, companies expect HMI solutions equipped with some Machine learning algorithms that can learn from all the data generated from IoT sensors and adapt to the ongoing behavior of the operator. However, there exist several challenges to implementing AI-powered HMIs, such as privacy concerns and the need for high-quality data to train AI models.

According to recent industry reports, the HMI (Human Machine Interface) market was valued at USD 4352.7 million in 2021~\cite{dhapte2023global}. The HMI market is predicted to witness significant growth, with an expected market size of USD 9652.6 million by the year 2030. This growth can be attributed to the increasing demand for advanced HMI solutions in a variety of industries, such as manufacturing, automotive, and healthcare, among others. The adoption of HMI solutions has been driven by the need for more intuitive and user-friendly interfaces that enhance productivity and efficiency while reducing errors and downtime. Additionally, the rapid advancements in technology, including AI and IoT, are expected to further boost the growth of the HMI market in the coming years. 

There are several companies that are currently using AI to create HMI solutions. The key players in the market include Siemens AG, Schneider Electric SE, ABB Ltd., and Rockwell Automation, Inc., Samsara. For example, Samsara offers an HMI that empowers operators with real-time asset health data and streamlines operations to enable a proactive approach to maintenance~\cite{samsara2020hmi}. Another company is Schneider Electric, which provides a comprehensive HMI solution that manages all vital machine features, including visualization, control, supervision, diagnostics, monitoring, and data logging. Schneider Electric's HMI solution is designed to be highly connectable and flexible, allowing for easy collaboration between operators and machines. Additionally, Rockwell Automation is another company that is using AI for creating HMI solutions. Rockwell Automation's HMI solution uses AI to provide predictive maintenance and real-time analytics to improve machine performance and reduce downtime.

Renesas offers a scalable SMARC 2.1 System on Module (SoM) with an AI solution based on the Renesas RZ/G2 microprocessor family and a full power and timing tree~\cite{renesas2023hmi}. This HMI SoM with AI accelerator is designed for AI-enabled human-machine interface (HMI) applications, such as smart industrial control terminals. The HMI SoM with AI accelerator is a turnkey architecture that provides a comprehensive solution for HMI applications, including visualization, control, supervision, diagnostics, monitoring, and data logging. Similarly, Microchip offers a range of smart human-machine interface (HMI) solutions that use machine learning to enhance the user experience and improve efficiency~\cite{microchip2023smart}. Their solutions include touch and touchless sensing, gesture recognition, and voice control. These technologies can be used in a variety of applications, including automotive, industrial, and medical devices. By leveraging machine learning, Microchip's HMI solutions can adapt to the user's behavior and preferences, enabling more intuitive and natural interactions with machines. Overall, Microchip's HMI solutions offer a reliable and innovative way to enhance the usability and efficiency of human-machine interfaces.

Technically Speaking, AI and one-shot learning can revolutionize the field of human-machine interaction by enabling more intuitive and natural interfaces that adapt to the user's behavior and preferences~\cite{frackiewicz2023future}. One-shot learning can enable machines to learn from a single example, reducing the need for extensive training data and making it easier for machines to adapt to new tasks and environments. The combination of AI and one-shot learning can enable machines to understand human emotions and intentions, enhancing the user experience and enabling more efficient and personalized interactions. The use of natural language processing (NLP) and voice recognition can enable more natural and intuitive interactions between humans and machines, reducing the need for physical interfaces and touchscreens. The integration of AI and one-shot learning in human-machine interaction can have significant implications for a wide range of industries, including healthcare, automotive, and manufacturing. 

To unleash the potential of AI for HMI solutions, edge AI techniques are to be adopted~\cite{soni2020edge}. In general, Edge AI brings intelligence closer to the devices, reducing the latency and improving the response time of HMIs. With edge AI, HMIs can adapt to the user's behavior and preferences, enabling more personalized and intuitive interactions.  Edge AI-powered HMIs can enable devices to understand and respond to voice commands, reducing the need for physical interfaces. Edge AI can enable HMIs to analyze data from sensors and other sources in real time, enabling predictive maintenance and improving the overall reliability of devices. Edge AI can enhance the security of devices by enabling them to detect and respond to threats in real-time, without relying on cloud-based solutions.

\section{AI for Milling Machines}\label{sec:milling}

In this section, we explain the milling machines and their components and applications. Afterward, we discuss the potential of leveraging AI technologies to develop efficient and safe milling machines. 

In general, a milling machine is a type of machinery used to remove material from a workpiece using rotary cutters~\cite{kumar2022milling}. The process of removing pieces of the material in line with the tool axis is known as milling, hence the name. Milling machines can drill, bore, and cut an array of materials and come in many types, used across various industries. There are several types of milling machines available, including computer-controlled vertical mills with the ability to move the spindle vertically along the Z-axis. Milling can be done with a wide range of machine tools, and it is one of the most commonly used processes for machining custom parts to precise tolerances. The original class of machine tools for milling was the milling machine (often called a mill). Disk- or barrel-shaped cutters are clamped through holes in their centers to arbors attached to the machine spindle; they have teeth on their peripheries only or on both peripheries and faces. An end mill is a cutter shaped like a pencil with a tapered shank that fits into the machine spindle; it has cutting teeth on its face and spiral blades on the lateral surface.

AI has several applications in milling machines. Specifically, it has the potential to enhance production capacity, boost machine productivity, and optimize the CNC (computer numerical control) machining industry by improving accuracy, reducing redundant tasks, predicting maintenance needs, optimizing parameters, and improving surface quality~\cite{soori2023machine}. Here are some ways in which AI and machine learning are being applied to CNC machining~\cite{hahn2022machine}:
\begin{itemize}
\item	Cutting forces and cutting tool wear prediction: Machine learning systems can be applied to predict cutting forces and cutting tool wear in CNC machine tools to increase cutting tool life during machining operations.
\item	Optimized machining parameters: Advanced machine learning systems can be used to obtain optimized machining parameters of CNC machining operations in order to increase efficiency during part manufacturing processes. For instance, such parameters can be tested within a digital twin —a virtual copy of the CNC machine—using machine learning and IoT to optimize CNC machinery.
\item	Surface quality prediction: Surface quality of machined components can be predicted and improved using advanced machine learning systems to improve the quality of machined parts.
\item	Machine scheduling: AI can be used for machine scheduling for optimized performance and downtime, and better data analysis, programming, and testing.
\item	Milling diagnosis: AI approaches can be used for milling diagnosis in composite sandwich structures based on honeycomb core.
\item	Reduction of redundant tasks: AI can reduce redundant tasks for manufacturers, which significantly impacts the time required for machining and human effort and makes the parts more identical to one another.
\end{itemize}

The utilization of modern generative models, e.g., OpenAI GPT4 and Google BARD, has the potential to revolutionize CNC programming by generating highly optimized and efficient programs~\cite{aloyan2023ai}. With their advanced natural language processing capabilities, the generative models are capable of comprehending intricate machining specifications and producing G and M codes tailored precisely to meet project requirements. This groundbreaking AI solution significantly saves time and effort during the programming process while minimizing the potential for errors inherent in manual programming methods. By incorporating generative models into CNC programming workflows, manufacturers can benefit from the following advantages:

\begin{itemize}
\item Accelerated program generation: the generative models can rapidly generate CNC programs, enabling manufacturers to expedite production timelines and meet demanding deadlines.
\item	Improved program quality: Leveraging AI-driven algorithms, the language model analyzes historical data to generate optimized programs that enhance machining efficiency and diminish tool wear.
\item Enhanced adaptability: the generative models' advanced algorithms enable it to adapt to evolving machining demands, ensuring that generated programs remain up-to-date and aligned with the latest requirements.
\item	Reduced reliance on skilled programmers: By minimizing the need for extensive human intervention, the generative models broadly reduce manufacturers' reliance on highly skilled CNC programmers, allowing for more efficient allocation of resources.
\end{itemize}

In general, machine learning models for CNC machining can be trained using various types of data, depending on the specific task or objective. Common types of data used for training machine learning models in CNC machining include:

\begin{itemize}
\item	Sensor data: CNC machines often have sensors that collect data during machining operations. This data can include information such as cutting forces, spindle speeds, tool temperatures, vibrations, and power consumption. Sensor data provides valuable insights into the machining process and can be used to train models for tasks like anomaly detection, fault diagnosis, or process optimization.
\item	Image data: Images captured by cameras or sensors placed within or around the CNC machine can be used to train models for tasks such as object detection, defect identification, or quality control. These images can provide visual information about the workpiece, tooling, or any abnormalities during the machining process.
\item	Time series data: CNC machining operations often generate time-stamped data that captures the changes and dynamics over time. Examples include sensor measurements sampled at regular intervals or historical process data. Time series data can be used to train models for tasks like forecasting, process control, or pattern recognition.
\item	CAD/CAM data: Computer-Aided Design (CAD) and Computer-Aided Manufacturing (CAM) data can be used to train models for tasks related to machining toolpath optimization, tool selection, or simulation. CAD/CAM data provides geometric information about the workpiece, tooling, and machining operations, enabling the models to learn from simulated or historical machining processes.
\item	 Historical production data: Historical data from past CNC machining operations, such as machining parameters, workpiece characteristics, and quality metrics, can be used to train models for tasks like predictive maintenance, process optimization, or estimating machining time.
\item	 Expert knowledge or rules: In addition to data-driven approaches, expert knowledge or domain-specific rules can be incorporated into machine learning models. These rules can be derived from expert machinists' experience or best practices in CNC machining.
\end{itemize}

The market for milling machines is projected to grow significantly in the coming years, with the adoption of predictive analytics driving growth~\cite{marketwatch2023micro}. The global market for milling machines is projected to reach \$10.6 billion by 2030, growing at a CAGR of 5.6\% over the analysis period 2022-2030. The market growth will be aided by the rising demand for customized mass production, as the milling machine produces customized products in large quantities. The micro-milling machine market has seen significant growth in recent years, driven by increased demand for its products and services. The integration of additive manufacturing and artificial intelligence is propelling the market growth of micro-milling machines. 

There are several companies that develop artificial intelligence CNC milling machines, including DMG Mori, TRUMPF, Mazak, ESPRIT, and Okuma~\cite{aloyan2023ai}. For instance, DMG Mori, a renowned machine tool manufacturer, is at the forefront of integrating AI systems into their CNC machines. The company is dedicated to enhancing machine performance, boosting productivity, and minimizing errors in the manufacturing process through their AI-powered solutions. DMG Mori's AI-driven technologies encompass intelligent monitoring systems that optimize the machining process. Furthermore, their advanced AI algorithms automatically adapt cutting parameters to minimize tool wear and prolong the lifespan of tools.

Mazak, another prominent player in the CNC industry, is actively driving the transformation of manufacturing through the development of AI-powered technologies. One notable innovation from Mazak is the introduction of AI-assisted programming features. These features enable their machines to learn from historical data, leveraging AI algorithms to generate optimized machining strategies. Mazak's AI-enhanced systems also incorporate real-time process monitoring, allowing for the early detection of potential issues. This proactive approach significantly reduces downtime and contributes to improved overall productivity. In addition, Mazak's Mazatrol SmoothAi system stands out for its ability to generate tool paths directly from the Mazatrol controller, requiring minimal human interaction, particularly when working with solid models.

Similarly, ESPRIT, a CNC machining software, is leveraging artificial intelligence (AI) to enhance its capabilities~\cite{lancaster2021cnc}. ESPRIT utilizes AI algorithms to optimize machining processes and improve productivity in the manufacturing industry. The software employs machine learning to analyze and adapt to various machining scenarios, enabling it to make intelligent decisions and optimize tool paths. By leveraging AI, ESPRIT aims to reduce cycle times, enhance tool life, and improve surface finish quality. Additionally, there exist specific AI features in ESPRIT, such as Adaptive Spindle Speed Control and AI-Driven Collision Avoidance, which help improve machining efficiency and prevent tool damage.

\section{Generative AI for Manufacturing}\label{sec:genai}

In this section, we discuss the adoption of generative AI (GenAI), such as Large Language Models (LLMs) and Generative Adversarial Networks (GANs), to optimize the manufacturing process. According to a recent KPMG study, conducted in March 2024 with over 280 decision-makers, 53\% of companies plan to increase their investments in GenAI within the next 12 months, with half of these companies aiming for a 40\% or more increase \cite{kpmg2024german}. Additionally, 67\% of companies anticipate a rise in turnover and automation, 65\% expect reduced costs, and 55\% hope for higher productivity. However, 37\% of respondents fear potential negative effects, such as job losses, data protection issues, and lack of employee acceptance. The study also highlights a need for improvement in strategy, governance, and training, as only about one-third of companies have a strategy in place, only 8\% have a comprehensive governance model, and 63\% do not feel adequately prepared for the EU's AI Act. Furthermore, only 38\% of respondents believe their company is well or very well prepared to train employees in handling GenAI-based solutions.

The manufacturing industry is currently grappling with numerous challenges, including cost pressures, supply chain disruptions, and the need to continuously innovate while maintaining safety and stability. This sector is also under significant pressure to adopt sustainable and eco-friendly practices amidst the global push for energy transition and reduced carbon footprints. GenAI is emerging as a powerful solution to many of these challenges. As highlighted in McKinsey's 2023 AI report, GenAI offers various capabilities that are gaining traction in the manufacturing field~\cite{Chui2023GenerativeAI}. For instance, \textit{summarization} is a key application where GenAI helps to improve the accessibility and usability of vast amounts of data stored in documents such as manuals and logbooks. For instance, IBM developed GenAI accelerators specifically for manufacturing to aid in this aspect, enhancing the efficiency with which operators and engineers access and utilize information~\cite{roy2024generative}. LLMs can also improve the management of complex documentation by efficiently organizing and parsing extensive texts, which facilitates quicker data retrieval. They help in drafting, editing, and validating detailed documents, ensuring accuracy and adherence to guidelines, which is crucial in technical and specialized fields.

Manufacturing firms often struggle with siloed and inconsistent data systems. GenAI can play a crucial role in creating a unified knowledge graph that provides \textit{contextual understanding} and accelerates data-driven decisions. IBM’s AI-powered Knowledge Discovery system is an example of such an application, helping to streamline data analysis and feature engineering in industrial settings. In \textit{software engineering}, LLMs contribute to automated code generation, code review, and bug identification, which streamlines development processes. They assist in maintaining up-to-date technical documentation and enable rapid prototyping and feature refinement. Tools like Microsoft Copilot and Watson Code Assistant utilize GenAI to facilitate faster and more efficient code development, thereby reducing manual labor and improving deployment times.

LLMs enable \textit{marketing} teams to create rich, original content that resonates with the brand’s voice across multiple platforms. These models personalize interactions, drive engagement, and improve conversion by analyzing consumer data. They also provide insights into market trends and consumer behavior and offer quick, accurate translations to help brands achieve global reach. Aside from marketing, in \textit{sales}, LLMs automate interactions and personalize communication This can drastically boost the efficiency of the sales process. LLMs help in lead generation, and content creation (such as emails and proposals), and integrate with CRM systems to enhance data management and follow-ups. Additionally, LLMs identify upselling and cross-selling opportunities, assist in sales training, and support advanced sales strategies like SEO-optimized product descriptions and automated contract management. LLMs transform \textit{customer service} by powering chatbots and virtual assistants that provide immediate, accurate, and personalized responses. They handle multiple interactions across different platforms, enhancing the overall customer experience while freeing human agents to solve more complex issues. 

Aside from customer service, LLMs can be a handy tool for \textit{product/service development}. LLMs use extensive market research and consumer insights to aid in product and service development. Makatura et al.~\cite{Makatura2024Large} assessed the capabilities of LLMs across various tasks, including design generation from natural language specifications; design space creation and variation; preparation of designs for manufacturing; design performance evaluation; and discovery of high-performing designs based on specific metrics. The authors highlighted the ability of LLMs to generate designs from high-level textual input and handle diverse representations and problem domains. LLMs demonstrate proficiency in creating parametric designs, suggesting parameter bounds, and generating design variations within a given design space. They also discussed the applications of LLMs in design for manufacturing (DfM), such as identifying suitable manufacturing techniques, suggesting design modifications, and creating manufacturing instructions.

Furthermore, they explored the potential of LLMs in evaluating designs based on objective, semi-subjective, and subjective criteria. LLMs can assist in inverse design by generating cost functions and optimized designs based on constraints. In \textit{asset management}, GenAI can develop foundation models for predictive maintenance, especially when failure data is scarce. These models can be pre-trained without labels and fine-tuned as needed, offering a robust method for maintaining equipment and predicting failures. Despite the potential of LLMs to streamline and enhance various aspects of the design and manufacturing process, There exist several limitations of LLMs in design and manufacturing. These limitations include challenges in analytical reasoning, complex computations, correctness, verification, and scalability. To address these limitations, several potential solutions such as using domain-specific languages, leveraging APIs for complex computations, implementing automated verification, decomposing large tasks, and adopting incremental design approaches, can be investigated and adopted.

\section{Conclusion}\label{sec:conclusion}
In this arcticle, we thoroughly explored the integration and impact of AI within the German manufacturing sector, offering a market analysis supported by current data and forecasting future trends. We have seen how AI technologies, particularly in the realms of predictive maintenance, process optimization, and safety enhancements, are revolutionizing the industry, allowing for increased operational efficiency and a competitive edge in a rapidly evolving market. The discussion highlighted the critical role of HMIs and presented a forward-looking perspective on the promising applications of GenAI and LLMs, which are poised to further transform product design, customer support, and knowledge management within manufacturing. Despite the challenges related to technology adoption, such as data quality and integration complexities, the potential for significant economic and operational benefits remains high.

As we look to the future, it is clear that the continued exploration and adoption of AI technologies will be crucial for the sustainability and growth of the manufacturing industry. The insights presented in this article should serve as a valuable resource for manufacturers, policymakers, and technology providers who are aiming to understand and leverage AI/GenAI to innovate and thrive in the digital age. The journey towards a fully AI-integrated manufacturing environment is complex and challenging, but as demonstrated, it is also rich with opportunities for transformation and advancement.

\balance
\bibliographystyle{ACM-Reference-Format}
\bibliography{main}

\end{document}